\documentclass[10pt,twocolumn,letterpaper]{article}

\usepackage[pagenumbers]{cvpr} % To force page numbers, e.g. for an arXiv version

\usepackage{graphicx}
\usepackage{amsmath}
\usepackage{amssymb}
\usepackage{booktabs}
\usepackage{algorithm}
\usepackage{algorithmic}
\usepackage{adjustbox}
\usepackage{cuted}

\usepackage[pagebackref,breaklinks,colorlinks]{hyperref}

\usepackage[capitalize]{cleveref}
\crefname{section}{Sec.}{Secs.}
\Crefname{section}{Section}{Sections}
\Crefname{table}{Table}{Tables}
\crefname{table}{Tab.}{Tabs.}

\def\cvprPaperID{*****} % *** Enter the CVPR Paper ID here
\def\confName{CVPR}
\def\confYear{2022}

\begin{document}

%\title{\LaTeX\ Author Guidelines for \confName~Proceedings}

%%%%%%%%% TITLE
\title{Enhanced Behavioral Cloning with Environmental Losses for Self-Driving Vehicles\vspace{-3ex}}

\author{Nelson Fernandez Pinto\\
Renault group\\
Paris, France\\
{\tt\small nelson.fernandez-pinto@renault.com\vspace{-9ex}}
\and
Thomas Gilles\\
Renault group\\
Paris, France\\
{\tt\small thomas.gilles@renault.com\vspace{-9ex}}
}
\maketitle

\begin{strip}\centering
\includegraphics[width=\textwidth]{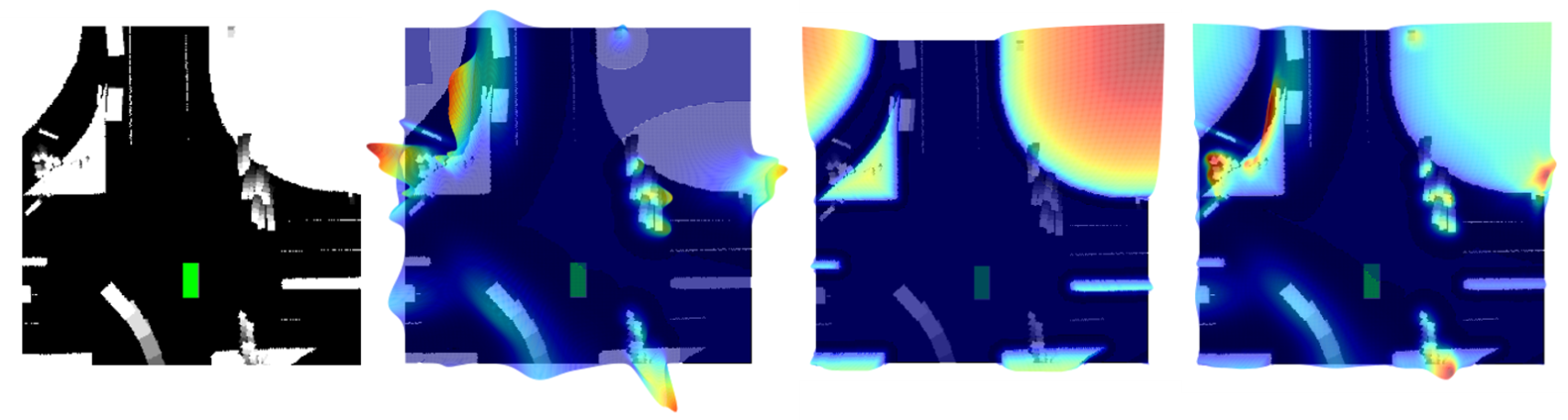}
\vspace*{-6ex}
\captionof{figure}{From left to right: top-down raster map of driving scene (ego vehicle in green), social loss function landscape, road loss function landscape, environmental loss function (superposition of social loss and road loss).
\label{fig:teaser}}
\end{strip}

%%%%%%%%% ABSTRACT
\begin{abstract}%\vspace{-1ex}
Learned path planners have attracted research interest due to their ability to model human driving behavior and rapid inference. Recent works on behavioral cloning show that simple imitation of expert observations is not sufficient to handle complex driving scenarios. Besides, predictions that land outside drivable areas can lead to potentially dangerous situations. This paper proposes a set of loss functions, namely Social loss and Road loss, which account for modelling risky social interactions in path planning. These losses act as a repulsive scalar field that surrounds non-drivable areas. Predictions that land near these regions incur in a higher training cost, which is minimized using backpropagation. This methodology provides additional environment feedback to the traditional supervised learning set up. We validated this approach on a large-scale urban driving dataset. The results show the agent learns to imitate human driving while exhibiting better safety metrics. Furthermore, the proposed methodology has positive effects on inference without the need to artificially generate unsafe driving examples. The explanability study suggests that the benefits obtained are associated with a higher relevance of non-drivable areas in the agent's decisions compared to classical behavioral cloning.

\end{abstract}

%%%%%%%%% BODY TEXT
\section{Introduction}

Self-driving vehicles (SDV) have the potential to establish a safer commute environment, removing human error, and dramatically decreasing the number of road fatalities in the coming years. Recently published data shows that semi-autonomous vehicles have a lower accident rate per mile than regular human drivers \cite{teslareport}. Although this indicator is promising and keeps improving over time, the rare occurrences of fatal accidents have a powerful effect on the public's opinion about the safety of SDV's \cite{kyriakidis2015public}. Therefore, there is an increasing pressure on manufacturers and regulators to improve the reliability of SDV perception and planning systems. Regardless of the progress achieved to date, there are still major challenges ahead in autonomous car technology. 

Path planning in complex and dynamic urban environments is one key aspect of autonomy and remains a relevant issue in the research community \cite{paden2016survey, torabi2019recent}. Model-based planners are widely used in robot and SDV navigation. However, these methods are complex to solve in real-time and often sub-optimal in multi-agent urban scenarios \cite{hadfield2017inverse, chen2018foad}. On the other hand, learned planners are able to model fine human driving behavior with low inference time. Among these techniques, behavioral cloning has attracted special interest due to its straight-forward training and implementation. This method consists of direct steering wheel angle regression from front-facing camera images using supervised learning \cite{bojarski2016end, torabi2019recent}. 

Recent works show that behavioral cloning exhibits poor performance in multi-agent situations and diminishing learning returns on large datasets \cite{codevilla2019exploring, bansal2018chauffeurnet}. Unsafe path planner predictions can lead to potentially catastrophic consequences \cite{calabresi2008cost, hecker2018failure}. Moreover, the presence of other vehicles, pedestrians and sidewalks increases the probability and extent of potential accidents.% Frequently, these situations are characterised as important divergences between the model's prediction and human driver manoeuvres \cite{hecker2018failure}. 

One of the biggest limitations of behavioral cloning is its inability to learn from interactions between the agent and the environment \cite{torabi2019recent}. Simple imitation of expert trajectories does not teach the agent to avoid undesired situations, such as colliding with other vehicles or driving outside the road. \cite{bansal2018chauffeurnet} proposed synthetic generation of unsafe driving examples, and its explicit penalisation using custom loss functions. The authors assigned a derivable cost to overlaps between predicted ego bounding boxes and non-drivable areas during training. Nevertheless, the proposed loss functions are not suitable for regression problems, a long-used behavioural cloning approach \cite{torabi2019recent,bojarski2016end}. Besides, finding the right proportion and realism of synthetic unsafe driving examples can be a challenging task.

In this paper we propose a set of training loss functions that increase an agent's awareness of regions to avoid while driving:  \emph{Social loss}, \emph{road loss} and their superposition \emph{environmental loss} (see figure \ref{fig:teaser}). These losses penalize excessive approach to non-drivable areas using heuristics during training. This penalization accounts for an additional environment feedback to the classic supervised learning set up. This method has measurable benefits over standard behavioral cloning, such as fewer collisions and overlaps with non-drivable areas. In addition, permanent effects are achieved during inference without needing to generate unsafe driving training examples. We validated this approach on a large-scale urban driving data set with offline safety metrics.

% Don't forget to cite figure 1

\subsection{Related work}
The earliest attempts of behavioural cloning go as far back as the 1980s with ALVINN \cite{pomerleau1989alvinn} using a shallow neural network to teach a car to follow the road from raw sensor inputs. The development of convolutional neural networks \cite{lecun1998gradient} and the recent progress of deep learning in computer vision \cite{krizhevsky2012imagenet, simonyan2014very} led to a resurgence of such end-to-end driving approach. In recent years, PilotNet \cite{bojarski2016end} used supervised learning to regress steering wheel angle command directly from front-facing images, without further interactions between the agent and the environment. The driving agent was able to keep road lanes in simple driving situations in highways. The following work from \cite{bojarski2017explaining} highlighted PilotNet's ability to learn relevant features such as road lanes, sidewalks and the presence of other vehicles from images. \cite{bansal2018chauffeurnet, djuric2018motion, cui2019multimodal, Hong_2019_CVPR} used the mediated perception approach, taking a top-down rasterized map as input to predict future ego positions trained with observational data.

\cite{codevilla2019exploring} trained a ConvNet in an end-to-end manner on both simulated and real-world data to perform path planning. This work highlighted the limited capabilities of behavioral cloning to handle complex urban scenarios. \cite{hecker2018failure} trained a ConvNet to predict divergences between model predictions and human driving. These divergences are associated with unsafe driving behavior, such as driving outside the road and colliding with other actors. 

\cite{chen2015deepdriving} used a raster map and ConvNets to predict affordance indicators, such as distance to road lanes, surrounding cars and differences between ego heading angle and road tangent. These indicators were mapped to steering commands using model-based equations. \cite{chen2019deep} added a model-based control module that actuates after the ConvNet's decision to avoid divergences from physically feasible trajectories during inference.
\cite{Ferguson2008EfficientlyUC} proposed a 2D cost occupancy grid generated in real-time using the perception stack. Each grid position is assigned a discrete cost varying from free to lethal. The path planning module uses heuristics to find the optimal trajectory. \cite{ferrer2015multi} proposes a multi-objective custom cost function to encourage a robot to select the shortest and least risky path during navigation. Nevertheless, the aforementioned approaches only address enhancing path planning safety during inference and not during training.

\cite{helbing1995social} proposed a model for pedestrian dynamics based on human social behavior. This model states that trajectories are chosen taking the shortest path while minimizing risky social interactions. This is because pedestrians feel increasingly uncomfortable as they get closer to a strange person. Besides, pedestrians also want to keep a safe distance to the border of buildings, streets and obstacles. This repulsive effect is modelled as a custom-designed monotonically decreasing vector field that accounts for the free space required to manoeuvre. Still, this method is not intended for vehicle path planning and is entirely model-based: It does not have learned parameters and does not imitate dynamic behavior from observations.

Custom design of reward/cost functions is a common practice in reinforcement learning. \cite{henaff2019model, liang2018cirl} used the overlap between ego and other vehicles, road lanes and sidewalks as a derivable cost to train a neural network in a simulated environment. \cite{bansal2018chauffeurnet} applied the same principle in supervised learning, generating unsafe driving examples and penalising them explicitly during training. Even so, the above-mentioned loss functions are suitable for 2D occupancy grids and not for regression problems. Furthermore, they only act when the unwanted overlap occurs, hence the need for examples of crashes in the data set. The loss functions we propose penalize proximity, so the agent gets feedback from the environment without the need for explicit overlaps.

%-------------------------------------------------------------------------
\section{Behavioral cloning with environmental losses}

Unlike \emph{classical} behavioral cloning, we use top-down raster maps of the driving scene as a replacement of front-facing camera images. Furthermore, instead of directly regressing steering wheel angle, we predict vehicle trajectory as a vector in the coordinate space using a ConvNet. Direct mapping between predicted trajectories and steering commands is possible with the SDV physical model. The agent is trained with supervised learning, using pairs of 3-channel 400x400 rasterized driving scenes maps and output target vectors of dimension 2x6. Each target vector consists of six future \emph{x,y} ego-centric coordinates sampled at 2 Hz, accounting for a 3-second prediction.

As mentioned in the introduction, classic behavioral cloning does not account for interactions between the agent and the environment. To overcome that, we modify the training loop, such that predicted trajectories are projected on the HD map crop. Then, we apply some heuristics to assign a derivable cost to proximity to non-drivable areas (traffic actors and sidewalks). The loss functions are designed in such a way that values are high in forbidden regions and decrease in function of distance. This cost is minimized during trained using backpropagation. The final train loss is the summation of the classic behavioral cloning imitation loss and the proposed heuristics.

\subsection{Building a raster map}
\begin{figure}[h]
  \centering
  \includegraphics[width=0.5\textwidth]{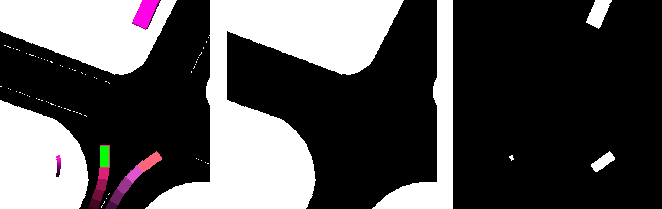}
  \caption{From left to right: Example of top-down raster map (ego vehicle in green), \textit{road layer} and \textit{traffic layer}.}
  \label{fig:layers}
\end{figure}

To build a top-down raster map we crop an HD map section of 30 by 30 meters around the SDV. The tile is built in such a way that the ego vehicle is located at the lower central part with 20 meters of forward vision pointing up.
Using the SDV's perception stack, we estimate the dimensions, heading angle, position and velocity of surrounding actors, such that pedestrians, vehicles and motorcycles. Then, we draw the traffic actor's bounding boxes on the top of the HD map crop, resulting in a synthetic representation of the driving scene. To exploit spatiotemporal information we applied two strategies: Encode actor's current speed as a RGB color, and draw the 6 past bounding boxes with a fading effect. The main motivation is to provide meaningful visual features that can be easily learned by a ConvNet. Unlike feature concatenation (between images and state vectors for example), the information is spatially linked to each actor during the 2D-convolution.
The velocity encoding was performed respecting the following pixel assignation:
\begin{itemize}
\item Channel R: 1 inside actor's bounding box, 0 otherwise.
\item Channel G: velocity module normalized from [0, 1].
\item Channel B: actor's heading angle normalized with zero mean unit variance. 
\end{itemize}
The fading effect was obtained by multiplying actor's previous bounding boxes by a linearly decreasing attenuation factor $\delta$ ranging from 1 (current position) to 1/6 in the last previous position.
Additionally, we output two semantic layers: \emph{Road layer} containing only drivable/non-drivable areas and \emph{traffic layer} with current positions of traffic actors. Figure  \ref{fig:layers} shows one example of a top-down raster map and its respective semantic layers. 

\subsection{Loss functions}

\subsubsection{Behavioral cloning imitation loss}
We used the Mean Square Error MSE loss function to penalise predictions that diverge from expert's trajectories. The MSE is widely used in imitation learning to teach agents to clone human driving behavior. Assuming we have a set of expert's trajectories $Y$ and model predictions $\hat{Y}$ in a time horizon $H$, the MSE is calculated as the average of the squared distance between target and predicted trajectories according to equation (1). 

\begin{equation}
MSE = \frac{\sum_{k=0}^{H-1}(Y_{to+k}-\hat{Y}_{to+k})^{2}}{H}
\end{equation}

%An early model of pedestrian dynamics stated that humans tend to navigate their environment avoiding social interactions with unknown people. Furthermore, pedestrians are also prone to keep a safe distance from the borders of buildings, streets and obstacles. A way to model this behavior is to suppose the existence of a monotonically decreasing repulsive field around non-transitable areas. In robot navigation, this field can be interpreted as a measure of the cost of choosing a given trajectory that we ideally would want to minimize.

\subsubsection{Social loss}
The social loss function expresses the idea that the agent should stay away from regions that are already occupied. The objective is to avoid collisions explicitly penalising excessive approaching to traffic actors. We model this concept by fitting 2D oriented gaussians at the top of each actor's bounding box. The value of the function evaluated in the ego pose is what we call Social Interaction (SI). SI reflects the magnitude of a \textit{repulsive} field that surrounds the actor's bounding box. It accounts for the free space the SDV needs to safely perform a maneuver. As its name indicates, SI only exists when another dynamic actor is on the scene, and it diminishes when it moves away from the ego vehicle.
 
The SI is calculated using the Gaussian ellipsoid formulation (2), where $(x_0, y_0)$, $\sigma_x$, $\sigma_y$ and angle $\theta $ are respectively the centroid, length, width and heading angle of a given actor's bounding box. A multiplicative constant \textit{K} allows us to set its maximum value, which can be related to the desired level of avoidance we assign to each actor.

\small
\begin{equation}
SI(x,y)=K\exp(-(a(x - x_0)^2 + 2b(x-x_0)(y-y_0) + c(y-y_0)^2))
\end{equation}
\normalsize

\textit{where:}

{\centering
$a = \frac{\cos^2\theta}{2\sigma_X^2} + \frac{\sin^2\theta}{2\sigma_Y^2}$\par
}
\bigbreak

{\centering
$b = -\frac{\sin2\theta}{4\sigma_X^2} + \frac{\sin2\theta}{4\sigma_Y^2}$\par
}
\bigbreak

{\centering
$c = \frac{\sin^2\theta}{2\sigma_X^2} + \frac{\cos^2\theta}{2\sigma_Y^2}$\par
}

\begin{figure}[h]
  \centering
  \includegraphics[width=0.45\textwidth]{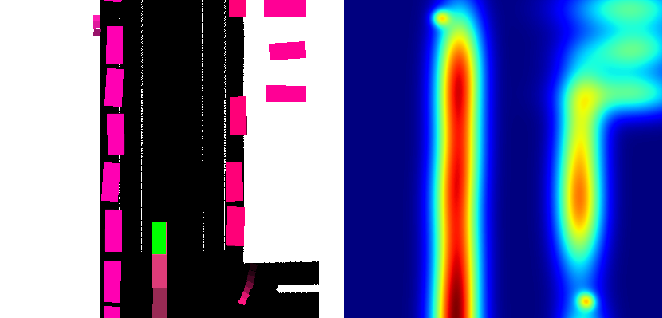}
  \caption{Example of top-down raster map (ego vehicle in green) and social loss landscape.}
  \label{fig:social_loss}
\end{figure}

\bigbreak
https://www.overleaf.com/project/5e6f5c815eb3d100018913f6
More generally, for a driving scenario with \textit{n} actors, the social loss is the superposition of individual social interactions evaluated over the current ego vehicle position according to equation (3). Figure \ref{fig:social_loss} shows an example of the social loss function landscape of a driving scene. Predictions that land near high intensity regions located above traffic actors incur in a higher training cost during optimization.    

\begin{equation}
 SocialLoss(x,y) = \sum_{i=1}^{n} K_iSI_i{(x_0,y_0, \theta, \sigma _{x}, \sigma _{y})}(x,y)
\end{equation}

\subsubsection{Road Loss}
The road loss function penalizes the agent when predictions land outside drivable areas. Ideally, we want a function whose intensity is low within the road and increases sharply when approaching the curbside. To approximate this behavior we use two derivable functions, which are computed if predictions are located inside or outside the road.

When the predictions are \emph {within} a drivable area, the road loss is calculated as a decreasing quadratic exponential function according to equation (4). This function is derivable and its decay can be easily controlled. After careful observation of the driver's habits, we choose the decay factor \emph{k} to obtain a 90\% attenuation at one meter from the nearest sidewalk.

\begin{equation}
DrivableLoss = \exp ^{-\frac{d^{2}}{k}}
\end{equation}
\quad \emph{Where \textit{d} is the euclidean distance to nearest non-drivable area.}

On the other hand, if the agent's predictions land outside the road, we penalise increasingly as they move further away from the drivable area. To do so, we use a logarithmic function with an added +1 to avoid discontinuity according equation (5). 

\begin{equation}
NonDrivableLoss = \log(d+1)
\end{equation}
\quad \emph{Where \textit{d} is the euclidean distance to nearest drivable area.}

\begin{figure}[H]
  \centering
  \includegraphics[width=0.45\textwidth]{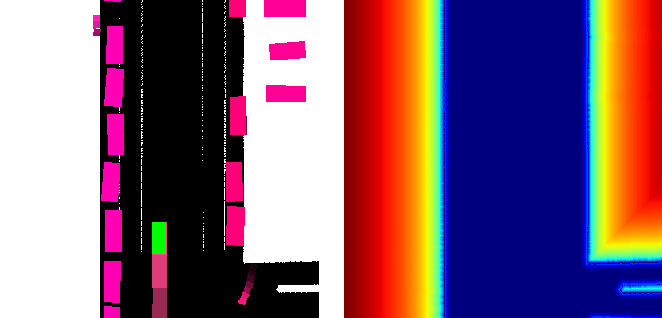}
  \caption{Example of top-down raster map (ego vehicle in green) and road loss landscape.}
  \label{fig:road_loss}
\end{figure}

The main particularity of the road loss function, is the need to estimate the distance between the predicted trajectory and the closest curbside at each forward pass. This means the function receives a trajectory vector and a \emph{road layer} image (see figure \ref{fig:layers}) as inputs. We designed algorithm (1) to locally approximate the road loss value. Figure \ref{fig:road_loss} shows one driving scene and its corresponding road loss landscape. As intended, the cost sharply increases when approaching curbsides and keeps increasing steadily as we move inside non-drivable areas.

\begin{algorithm}[H]
\caption{Algorithm for road loss computation}
\begin{algorithmic}[1]
\renewcommand{\algorithmicrequire}{\textbf{Input:}}
\renewcommand{\algorithmicensure}{\textbf{Output:}}
\REQUIRE Ego vehicle trajectory prediction $p_{t_0:t_{0+H}} = ((x_t,y_t)$ for t in $[t_0, t_0+H-1])$\\
Road layer M : 1 if pixel is non-drivable, else 0
\ENSURE  
\STATE \textit{Initialisation} : L=0
\FOR {$i = 0$ to $H-1$}
\STATE $p = p_{t_0+i}$
\IF {$p$ in on drivable area (M[p]=0)}
\STATE Compute L2 distance $d$ to closest non-drivable pixel
\STATE L += DrivableLoss   (4)
\ELSE
\STATE Compute L2 distance $d$ to closest drivable pixel
\STATE L += NonDrivableLoss   (5)
\ENDIF
\ENDFOR
\RETURN $L/H$
\end{algorithmic}
\end{algorithm}

\subsubsection{Enhanced behavioral cloning loss function}
The final loss function is the summation of the classic behavioral cloning MSE loss function and the environmental losses. The goal is that MSE will teach the network to imitate expert's trajectories while the environmental losses will provide useful contextual feedback directly from the environment. A multiplicative constant \emph{K} is added so we are able to control the influence of environmental losses in regards to the imitation loss. In short, for a pair of target and predicted trajectories $Y$ and $\widehat{Y}$, a tensor of actor's bounding boxes $TrafficBoundingBoxes$ and an HD map crop mask $Road Layer$, the loss function can be calculated using the following expression:

\small
\begin{eqnarray*}
Loss & = & MSE(Y, \widehat{Y}) \\
& & {} + K_1SocialLoss(TrafficBoundingBoxes,\widehat{Y})\\
& & {} + K_2RoadLoss(RoadLayer,\widehat{Y}) \\
\end{eqnarray*}
\normalsize

\begin{figure}[h]
  \centering
  \includegraphics[width=0.5\textwidth]{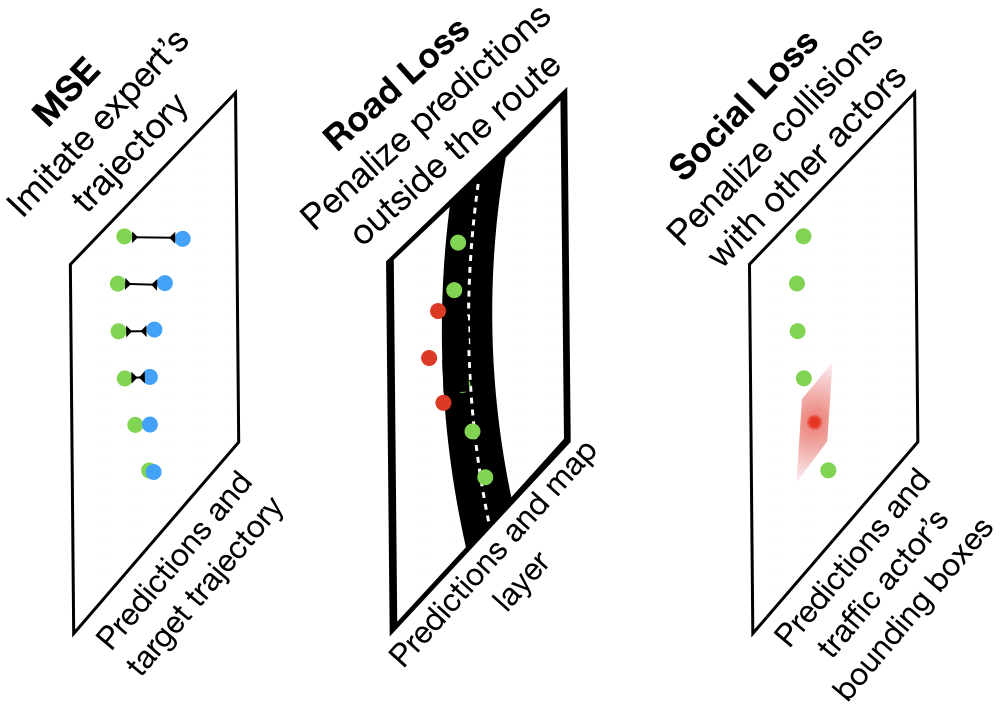}
  \caption{Representation of training supervisions.}
  \label{fig:supervisions}
\end{figure}

Figure \ref{fig:supervisions} shows a graphic overview of the three supervisions, their goals and inputs. 

\subsection{Network architecture}
\begin{figure}[h]
  \centering
  \includegraphics[width=0.5\textwidth]{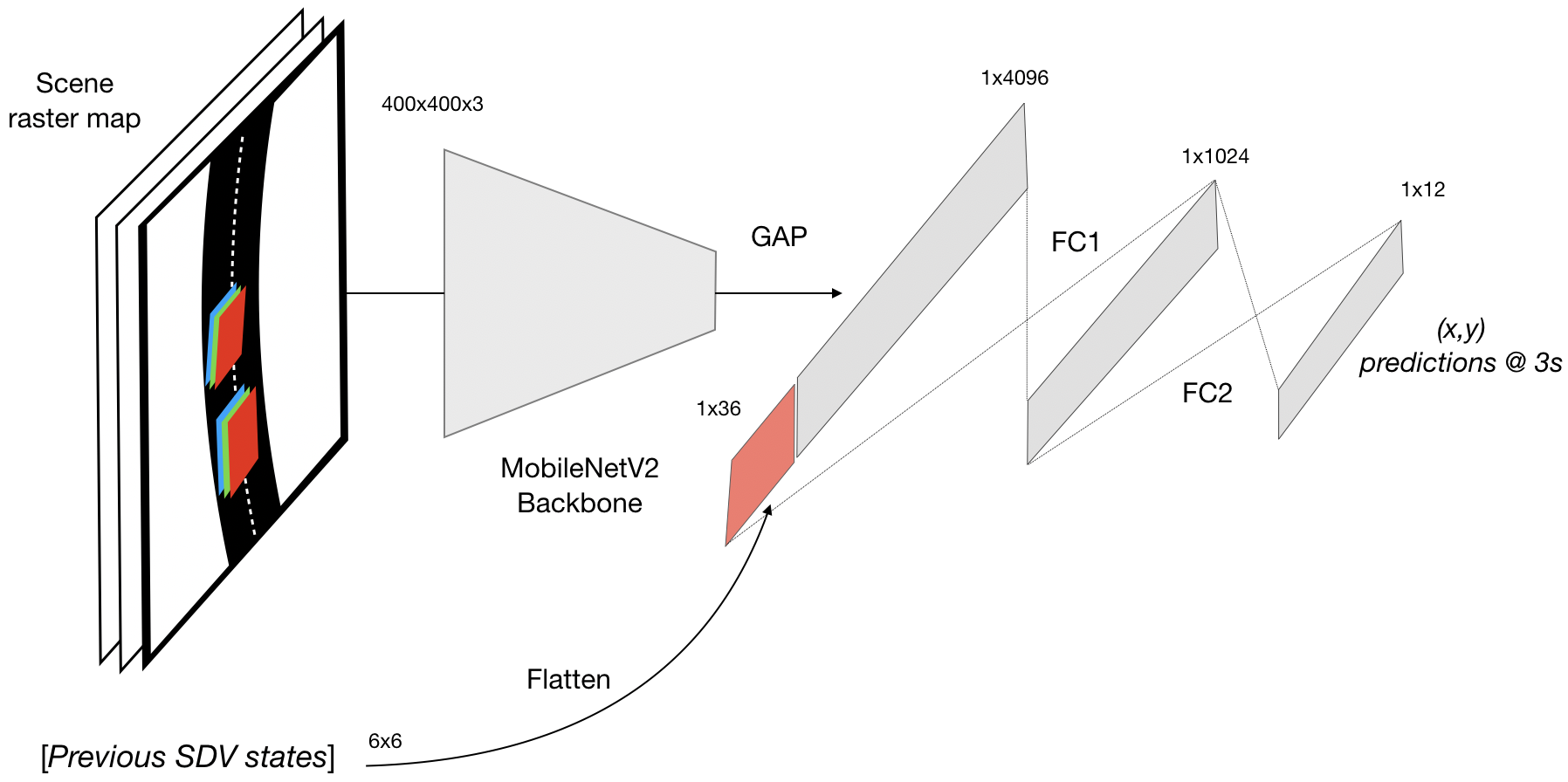}
  \caption{ConvNet raster map to trajectory regressor architecture.}
  \label{fig:architecture}
\end{figure}

We used a state-of-the-art MobilenetV2 architecture \cite{mobilenetv2} to perform image to trajectory regression. This architecture is widely known for being lightweight and having fast inference, making it suitable for real-time systems. We modified the fully connected layers to have an output size of dimension 1x12. This corresponds to 6 future \emph{x, y} ego vehicle positions. The model takes as input a 3-channel raster map of 400x400 pixels and a vector of past normalized ego states. Ego states include: Previous six \emph{x, y} positions, speed, acceleration, heading angle and heading angle speed. The convolutional backbone extracts features from the driving scene raster map, that are aggregated using Global Average Pooling (GPA). The state vector is then concatenated to the feature vector and feed into the fully connected regressor. The whole network is trained in a end-to-ed manner using backpropagation. Figure \ref{fig:architecture} shows the proposed architecture with inputs and output sizes.

\subsection{Validation strategy}
Using the network's predicted SDV trajectories $\hat{Y}$ and the vehicle dimensions (Renault Zoe 4.17x1.73m) we draw ego bounding boxes at the top of the HD map at each forward pass. Then, we compute the overlap between ego and non-drivable areas with the \emph{road layer} and \emph{traffic layer} binary masks (see Figure \ref{fig:layers}). The overlap is normalized to obtain two safety indexes: Collision index and out of road index.

The collision index is the average superposition of predicted bounding boxes and other actors in squared meters. It is calculated using the expression (6), where $H$ is the predicted time horizon, $res$ is the HD map resolution, $Ego$ and $obj$ are the respective predicted and traffic bounding boxes. 

\begin{equation}
COLLIndex = \frac{1}{H*res^{2}}\sum_{i=1, j=1}^{i=H, j=n}Ego(\hat{Y})_{i}  \cap  obj_{j}  [m^{2}]
\end{equation}

In an analogous way, we measure the average overlap of predicted SDV bounding boxes and sidewalks. We call this indicator Out of Road Index (OORIndex) which is calculated using expression (7).

\small
\begin{equation}
OORIndex = \frac{1}{H*res^{2}}\sum_{i=1}^{i=H}Ego(\hat{Y})_{i}  \cap  NonDrivableArea  [m^{2}]\\
\end{equation}
\normalsize

Also, we apply the guided back-propagation (GB) method \cite{GBP} to explain the agent's decisions during inference. Its working principle is to propagate gradients backwards from the output using deconvolution until reaching the input image. In consequence, the resulting heat map is the same size as the input frame. Highlighted regions are consistent with learned representations of deep layers of the convolutional neural network \cite{GBP}. Then, we calculate the normalised scalar product between the gradients heat map and the driving scene layers (see figure \ref{fig:explainability} and figure \ref{fig:layers}) according to expressions (8) and (9). Doing so, we quantify the agent's \emph{awareness} to specific objects, such as other vehicles and sidewalks. 

%In addition, we apply the guided back-propagation (GB) method \cite{GBP} to explain agent's decisions during inference. Its working principle is to propagate gradients backwards from the output using deconvolution until reaching the input image. In consequence, the resulting heat map is the same size as the input frame. Highlighted regions are consistent with learned representations of deep layers of the convolutional network. We use this information to estimate the agent's \emph{awareness} to specific objects, such as other vehicles and sidewalks. As index, we calculate the normalised scalar product between the gradient heat map and the driving scene layers (see Fig. \ref{fig:explainability} and Fig. \ref{fig:layers}) according to expressions (8) and (9).

\begin{equation}
SocialIndex =  \frac{GradientHeatMap\cdot TrafficLayer}{sum(GradientHeatMap)}   
\end{equation}

\begin{equation}
MapIndex =  \frac{GradientHeatMap\cdot RoadLayer}{sum(GradientHeatMap)}  
\end{equation}

Intuitively, if highlighted pixels are located on areas corresponding to vehicles and pedestrians, the social index will increase. Likewise, superpositions with sidewalks increase the map index.

\section{Dataset and experiments}
We apply the proposed methodology on a public large scale urban driving dataset. Nuscenes \cite{caesar2019nuscenes} is composed by 1000 dense traffic sequences of 20 seconds collected in Boston and Singapore. This dataset includes 1.4 million manually annotated bounding boxes, 390K LIDAR sweeps, HD map semantic layers and high fidelity GPS localisation. The driving sequences are complex, including multiple agents and intricate road layouts. We used the built-in Nuscenes train and validation partitions consisting of 700 and 150 sequences of 20-seconds respectively. We subdivided the sequences in 13 time contiguous time steps, including 6 previous, current and 6 future ego coordinates as targets. Such examples were generated using a sliding window of step 1 second, resulting in 19730 training and 4219 validation samples. The training was performed with a single NVIDIA RTX 6000 GPU with 24Gb of memory, Adam optimizer with learning rate 1e-3 and batch size of 16 elements.

\subsection{Experiment 1: Effect of environmental losses on MSE and safety indexes}

In the first experience, we studied the effect of adding the social loss term to the classic behavioral cloning MSE loss function. For that, we trained several models varying a multiplicative factor $K_1$ from 0 (only MSE) to 10 according to equation (10). We also calculated the OOR and COLL indexes of expert's demonstrations as reference.%For each model we computed the MSE and collision index.% We aim to verify if the social loss reduces the amount of collisions during inference and its effects on the MSE.

\begin{equation}
Loss = MSE + K_1SocialLoss
\end{equation}

\begin{figure}[h]
  \centering
  \includegraphics[width=0.5\textwidth]{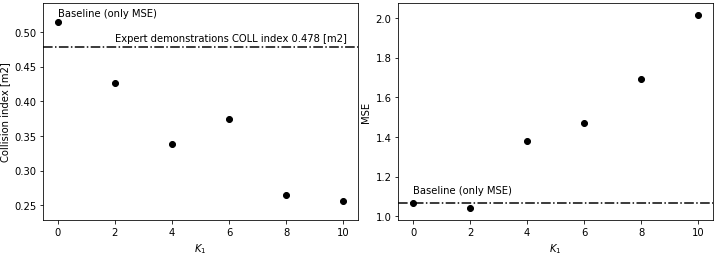}

  \caption{Social loss $K_1$ vs Collision index (left) and MSE (right).}
  \label{fig:study_social_loss}
\end{figure}

Figure \ref{fig:study_social_loss} left shows that when increasing the factor $K_1$ the collision index decreases consistently. After $K_1$ = $1$ all models exhibit a lower collision index than the expert's reference. On the other hand, the social loss becomes dominant and the MSE starts to increase (right). This result is expected as the agent is being encouraged to maintain a greater safety distance from other traffic actors. This distance can be even greater than that taken by the human driver in the training dataset. We select $K_1$ $=2$ to obtain a lower MSE and a lower collision index compared to the MSE-only baseline.

In a similar way, we repeat the previous experience using road loss according to equation (11).

\begin{equation}
Loss = MSE + K_2RoadLoss
\end{equation}

\begin{figure}[h]
  \centering
  \includegraphics[width=0.5\textwidth]{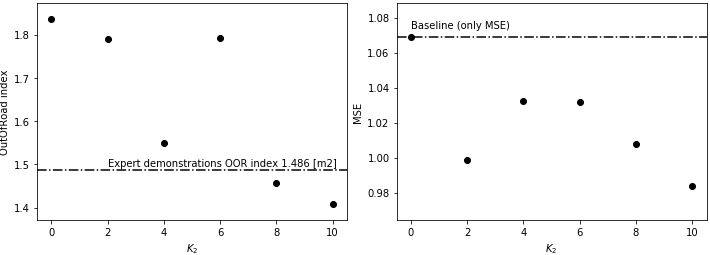}
  \caption{Road loss $k_2$ vs OOR index (left) and MSE (right)}
  \label{fig:study_map_loss}
\end{figure}

\small
\begin{table*}[h!]
\begin{adjustbox}{width=1\textwidth}
\label{table_example}
%\begin{center}
\begin{tabular}{|c||c||c||c||c||c||c||c|}
\hline
 Loss & MSE & COLLIndex[m2] & OORIndex[m2] & TotalOverlap[m2] & SocialIndex & MapIndex \\
\hline
 $Expert$ $demonstrations$ & N/A& 0.478& 1.486& 1.964& N/A& N/A\\

\hline
 $MSE (1)$ & 1.069& 0.520& 1.861& 2.382& 1.845& 0.508\\
\hline
 $MSE (1) + SocialLoss (3)$ & 1.045& 0.439& 1.882& 2.321& 2.414& 0.665\\
\hline
 $MSE (1) + RoadLoss (4)(5)$ & 0.983& 0.490& 1.456& 1.946& 1.878& \ 0.711\\
\hline
  $MSE (1) + EnvLoss (3)(4)(5)$ & 1.040& \textbf{0.479}& \textbf{1.440}& \textbf{1.919}& 2.324& 0.560\\
\hline
\end{tabular}
%\end{center}
\end{adjustbox}
\caption{Ablation study on the effect of environmental losses.}
\end{table*}
\normalsize

Figure \ref{fig:study_map_loss} left shows that when increasing the road loss intensity, the OOR index decreases, meaning fewer predicted trajectories land over non-drivable areas. However, the tendency is noisy  especially around $K_2$ = $6$. This large variance is associated with model initialization and the use of a stochastic solver. On the other hand, reductions on the MSE suggest cooperative interactions between losses.
%we obtain significant MSE reductions evidencing 
%evidencing better generalisation capabilities.  %Furthermore, after $K_2$ $=8$ the agent exhibits lower MSE than the baseline model and lower OORindex compared to the expert's demonstrations.

\subsection{Experiment 2: Ablation study}

%Once the optimal \emph{K} values of the environmental losses are set, 
We performed a full ablation study to measure the effect on the environmental losses during inference. The obtained MSE, safety and awareness indexes (see section 2.4) are reported in Table 1. Models trained with environmental losses exhibit lower MSE compared to the baseline showing improved generalization. In consistency with section 3.1, each environmental loss has a positive effect on reducing a safety metric: Social loss model shows fewer overlaps with traffic actors and the road loss model with non-drivable areas. When using both environmental losses, the model exhibits the lowest total overlap (COLLIndex + OORIndex) and therefore the most conservative driving behavior, similar to expert demonstrations. Increases of the awareness indexes, suggest that models trained with environmental losses learn internal representations that match non-drivable regions. 

\begin{figure}[ht]
  \centering
  \includegraphics[width=0.5\textwidth]{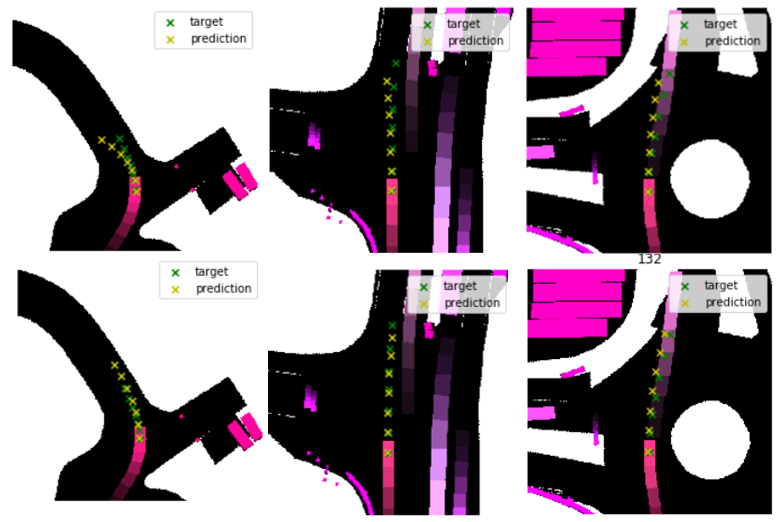}
  \caption{Prediction examples of baseline (top row) and environmental losses (bottom row) model.}
  \label{fig:driving_examples}
\end{figure}

Figure \ref{fig:driving_examples} shows some examples of test set driving scenarios and the respective model predictions. The model trained with environmental losses (bottom row) follows more closely the expert demonstrations (green crosses) compared to the MSE-only baseline (top row).

\begin{figure}[ht]
  \centering
  \includegraphics[width=0.5\textwidth]{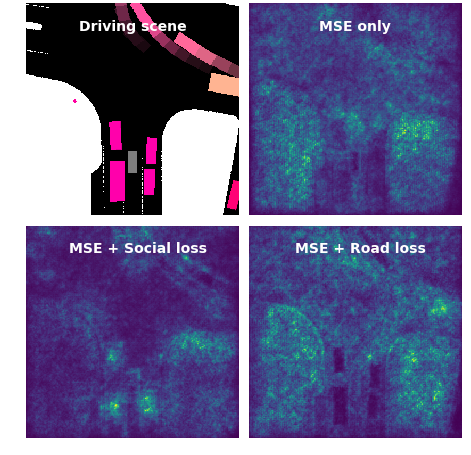}
  \caption{Driving scene and explanatory guided-gradient heat maps of trained models.}
  \label{fig:explainability}
\end{figure}

Figure \ref{fig:explainability} shows the explainability heat maps of a driving scene obtained with the trained models. The social loss model (bottom left) highlights with more precision vehicles around ego compared to baseline (top right). A similar behavior is obtained with the road loss model (bottom right) that focuses on curbsides. This \emph{watch this effect} is consistent with the increased Social and Map awareness indexes of table 1. % These results suggest an induced \emph{watch this} effect, as the network learns features that match non-drivable areas of the driving scene.
%effect due to the use of environmental losses during training. This means, the network is guided to discover features that match with non-drivable areas of the driving scene.

\section{Discussion}
We presented a set of training losses that allows to improve classic behavioral cloning by providing direct feedback from the environment. The aforementioned functions are derivable in regression problems and impact positively key safety metrics without the need to generate unsafe driving training examples. This is achieved by penalising proximity instead of hard overlaps with non-drivable areas. A set of adjustable parameters control the loss intensity to encourage a more conservative driving behavior during inference. The awareness indexes and explanatory maps suggest that these benefits are associated to an increased relevance of non-drivable areas compared to classical behavioral cloning. We consider the incorporation of temporal information directly into the losses as a promising improvement direction for following works.

{\small
\bibliographystyle{ieee_fullname}
\bibliography{egbib}

\begin{thebibliography}{10}\itemsep=-1pt

\bibitem{teslareport}
Tesla vehicle safety report.
\newblock \url{https://www.tesla.com/VehicleSafetyReport}.
\newblock Accessed: 2019-09-19.

\bibitem{bansal2018chauffeurnet}
Mayank Bansal, Alex Krizhevsky, and Abhijit Ogale.
\newblock Chauffeurnet: Learning to drive by imitating the best and
  synthesizing the worst.
\newblock {\em arXiv preprint arXiv:1812.03079}, 2018.

\bibitem{bojarski2016end}
Mariusz Bojarski, Davide Del~Testa, Daniel Dworakowski, Bernhard Firner, Beat
  Flepp, Prasoon Goyal, Lawrence~D Jackel, Mathew Monfort, Urs Muller, Jiakai
  Zhang, et~al.
\newblock End to end learning for self-driving cars.
\newblock {\em arXiv preprint arXiv:1604.07316}, 2016.

\bibitem{bojarski2017explaining}
Mariusz Bojarski, Philip Yeres, Anna Choromanska, Krzysztof Choromanski,
  Bernhard Firner, Lawrence Jackel, and Urs Muller.
\newblock Explaining how a deep neural network trained with end-to-end learning
  steers a car.
\newblock {\em arXiv preprint arXiv:1704.07911}, 2017.

\bibitem{caesar2019nuscenes}
Holger Caesar, Varun Bankiti, Alex~H Lang, Sourabh Vora, Venice~Erin Liong,
  Qiang Xu, Anush Krishnan, Yu Pan, Giancarlo Baldan, and Oscar Beijbom.
\newblock nuscenes: A multimodal dataset for autonomous driving.
\newblock {\em arXiv preprint arXiv:1903.11027}, 2019.

\bibitem{calabresi2008cost}
Guido Calabresi.
\newblock {\em The cost of accidents: a legal and economic analysis}.
\newblock Yale University Press, 2008.

\bibitem{chen2015deepdriving}
Chenyi Chen, Ari Seff, Alain Kornhauser, and Jianxiong Xiao.
\newblock Deepdriving: Learning affordance for direct perception in autonomous
  driving.
\newblock In {\em Proceedings of the IEEE International Conference on Computer
  Vision}, pages 2722--2730, 2015.

\bibitem{chen2018foad}
Jianyu Chen, Changliu Liu, and Masayoshi Tomizuka.
\newblock Foad: Fast optimization-based autonomous driving motion planner.
\newblock In {\em 2018 Annual American Control Conference (ACC)}, pages
  4725--4732. IEEE, 2018.

\bibitem{chen2019deep}
Jianyu Chen, Bodi Yuan, and Masayoshi Tomizuka.
\newblock Deep imitation learning for autonomous driving in generic urban
  scenarios with enhanced safety.
\newblock {\em arXiv preprint arXiv:1903.00640}, 2019.

\bibitem{codevilla2019exploring}
Felipe Codevilla, Eder Santana, Antonio~M L{\'o}pez, and Adrien Gaidon.
\newblock Exploring the limitations of behavior cloning for autonomous driving.
\newblock {\em arXiv preprint arXiv:1904.08980}, 2019.

\bibitem{cui2019multimodal}
Henggang Cui, Vladan Radosavljevic, Fang-Chieh Chou, Tsung-Han Lin, Thi Nguyen,
  Tzu-Kuo Huang, Jeff Schneider, and Nemanja Djuric.
\newblock Multimodal trajectory predictions for autonomous driving using deep
  convolutional networks.
\newblock In {\em 2019 International Conference on Robotics and Automation
  (ICRA)}, pages 2090--2096. IEEE, 2019.

\bibitem{djuric2018motion}
Nemanja Djuric, Vladan Radosavljevic, Henggang Cui, Thi Nguyen, Fang-Chieh
  Chou, Tsung-Han Lin, and Jeff Schneider.
\newblock Motion prediction of traffic actors for autonomous driving using deep
  convolutional networks.
\newblock {\em arXiv preprint arXiv:1808.05819}, 2018.

\bibitem{Ferguson2008EfficientlyUC}
Dave Ferguson and Maxim Likhachev.
\newblock Efficiently using cost maps for planning complex maneuvers.
\newblock 2008.

\bibitem{ferrer2015multi}
Gonzalo Ferrer and Alberto Sanfeliu.
\newblock Multi-objective cost-to-go functions on robot navigation in dynamic
  environments.
\newblock In {\em 2015 IEEE/RSJ International Conference on Intelligent Robots
  and Systems (IROS)}, pages 3824--3829. IEEE, 2015.

\bibitem{hadfield2017inverse}
Dylan Hadfield-Menell, Smitha Milli, Pieter Abbeel, Stuart~J Russell, and Anca
  Dragan.
\newblock Inverse reward design.
\newblock In {\em Advances in neural information processing systems}, pages
  6765--6774, 2017.

\bibitem{hecker2018failure}
Simon Hecker, Dengxin Dai, and Luc Van~Gool.
\newblock Failure prediction for autonomous driving.
\newblock In {\em 2018 IEEE Intelligent Vehicles Symposium (IV)}, pages
  1792--1799. IEEE, 2018.

\bibitem{helbing1995social}
Dirk Helbing and Peter Molnar.
\newblock Social force model for pedestrian dynamics.
\newblock {\em Physical review E}, 51(5):4282, 1995.

\bibitem{henaff2019model}
Mikael Henaff, Alfredo Canziani, and Yann LeCun.
\newblock Model-predictive policy learning with uncertainty regularization for
  driving in dense traffic.
\newblock {\em arXiv preprint arXiv:1901.02705}, 2019.

\bibitem{Hong_2019_CVPR}
Joey Hong, Benjamin Sapp, and James Philbin.
\newblock Rules of the road: Predicting driving behavior with a convolutional
  model of semantic interactions.
\newblock In {\em The IEEE Conference on Computer Vision and Pattern
  Recognition (CVPR)}, June 2019.

\bibitem{krizhevsky2012imagenet}
Alex Krizhevsky, Ilya Sutskever, and Geoffrey~E Hinton.
\newblock Imagenet classification with deep convolutional neural networks.
\newblock In {\em Advances in neural information processing systems}, pages
  1097--1105, 2012.

\bibitem{kyriakidis2015public}
Miltos Kyriakidis, Riender Happee, and Joost~CF de Winter.
\newblock Public opinion on automated driving: Results of an international
  questionnaire among 5000 respondents.
\newblock {\em Transportation research part F: traffic psychology and
  behaviour}, 32:127--140, 2015.

\bibitem{lecun1998gradient}
Yann LeCun, L{\'e}on Bottou, Yoshua Bengio, and Patrick Haffner.
\newblock Gradient-based learning applied to document recognition.
\newblock {\em Proceedings of the IEEE}, 86(11):2278--2324, 1998.

\bibitem{liang2018cirl}
Xiaodan Liang, Tairui Wang, Luona Yang, and Eric Xing.
\newblock Cirl: Controllable imitative reinforcement learning for vision-based
  self-driving.
\newblock In {\em Proceedings of the European Conference on Computer Vision
  (ECCV)}, pages 584--599, 2018.

\bibitem{paden2016survey}
Brian Paden, Michal {\v{C}}{\'a}p, Sze~Zheng Yong, Dmitry Yershov, and Emilio
  Frazzoli.
\newblock A survey of motion planning and control techniques for self-driving
  urban vehicles.
\newblock {\em IEEE Transactions on intelligent vehicles}, 1(1):33--55, 2016.

\bibitem{pomerleau1989alvinn}
Dean~A Pomerleau.
\newblock Alvinn: An autonomous land vehicle in a neural network.
\newblock In {\em Advances in neural information processing systems}, pages
  305--313, 1989.

\bibitem{mobilenetv2}
Mark Sandler, Andrew~G. Howard, Menglong Zhu, Andrey Zhmoginov, and
  Liang{-}Chieh Chen.
\newblock Inverted residuals and linear bottlenecks: Mobile networks for
  classification, detection and segmentation.
\newblock {\em CoRR}, abs/1801.04381, 2018.

\bibitem{simonyan2014very}
Karen Simonyan and Andrew Zisserman.
\newblock Very deep convolutional networks for large-scale image recognition.
\newblock {\em arXiv preprint arXiv:1409.1556}, 2014.

\bibitem{GBP}
Jost~Tobias Springenberg, Alexey Dosovitskiy, Thomas Brox, and Martin
  Riedmiller.
\newblock Striving for simplicity: The all convolutional net.
\newblock {\em arXiv preprint arXiv:1412.6806}, 2014.

\bibitem{torabi2019recent}
Faraz Torabi, Garrett Warnell, and Peter Stone.
\newblock Recent advances in imitation learning from observation.
\newblock {\em arXiv preprint arXiv:1905.13566}, 2019.

\end{thebibliography}
}

\end{document}